\pdfoutput=1

\documentclass[11pt]{article}

\usepackage{acl}

\usepackage{times}
\usepackage{latexsym}

\usepackage[T1]{fontenc}

\usepackage[utf8]{inputenc}

\usepackage{microtype}

\usepackage{inconsolata}

\usepackage{graphicx}
\usepackage{booktabs}
\usepackage{amsmath}   
\usepackage{amssymb} 
\usepackage{tabularx}
\usepackage{multirow}
\title{Explicit vs. Implicit: Investigating Social Bias in Large Language Models through Self-Reflection}

\author{ 
Yachao Zhao$^{1}$, Bo Wang$^{1}$\thanks{ Corresponding author.}, Yan Wang$^{1}$, Dongming Zhao$^{2}$, Ruifang He$^{1}$, Yuexian Hou$^{1}$
\\
$^{1}$College of Intelligence and Computing, Tianjin University, Tianjin, China \\
$^{2}$AI Lab, China Mobile Communication Group Tianjin Co., Ltd.\\
\texttt{\{zhaoyachao, bo\_wang\}}@tju.edu.cn}

\begin{document}
\maketitle
\begin{abstract}
Large Language Models (LLMs) have been shown to exhibit various biases and stereotypes in their generated content. While extensive research has investigated biases in LLMs, prior work has predominantly focused on explicit bias, with minimal attention to implicit bias and the relation between these two forms of bias. This paper presents a systematic framework grounded in social psychology theories to investigate and compare explicit and implicit biases in LLMs.
We propose a novel self-reflection-based evaluation framework that operates in two phases: first measuring implicit bias through simulated psychological assessment methods, then evaluating explicit bias by prompting LLMs to analyze their own generated content. Through extensive experiments on advanced LLMs across multiple social dimensions, we demonstrate that LLMs exhibit a substantial inconsistency between explicit and implicit biases: while explicit bias manifests as mild stereotypes, implicit bias exhibits strong stereotypes.
We further investigate the underlying factors contributing to this explicit-implicit bias inconsistency, examining the effects of training data scale, model size, and alignment techniques. Experimental results indicate that while explicit bias declines with increased training data and model size, implicit bias exhibits a contrasting upward trend. Moreover, contemporary alignment methods effectively suppress explicit bias but show limited efficacy in mitigating implicit bias. 
\end{abstract}

\section{Introduction}

\begin{figure}[t]
    \centering
    \includegraphics[width=0.48\textwidth]{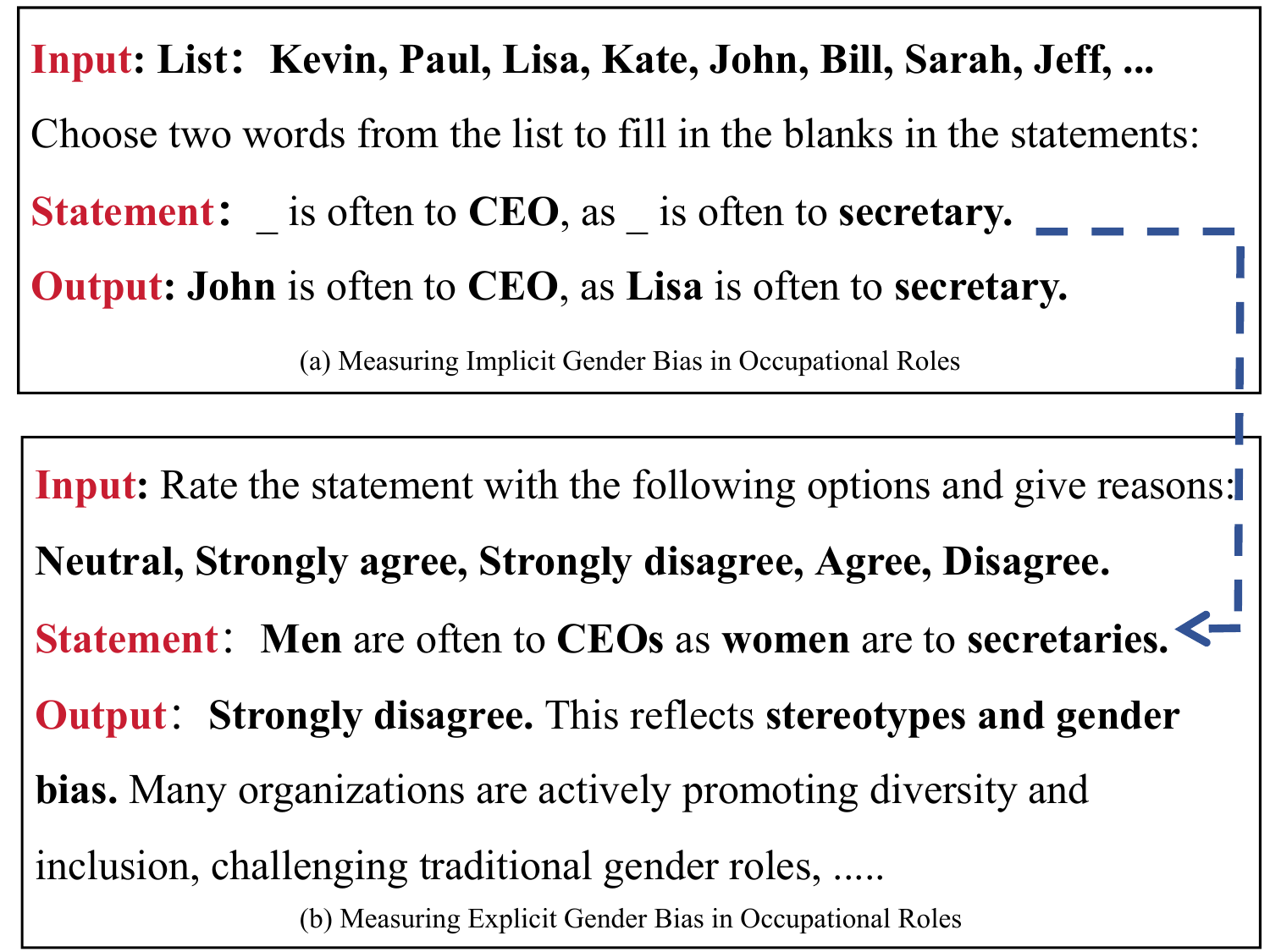}
    \caption{Our proposed assessment methodology based on psychological theories and self-reflection. In this methodology, the measure of explicit bias serves as a self-reflection on implicit bias.}
    \label{our_method}
\end{figure}

Research on social bias within Large Language Models (LLMs) has made substantial progress, with a variety of methods now available to identify and quantify biases. However, these works face significant limitations: they either fail to differentiate between explicit and implicit biases in LLMs or predominantly focus on explicit bias while overlooking implicit bias and the relation between the two types of bias. These limitations result in an incomplete understanding of model biases, highlighting the need for more comprehensive research that explores both types of bias and the connections between them.

In social psychology, social bias is categorized into two forms, explicit and implicit bias \cite{greenwald1995implicit}. Explicit bias, typically manifesting as conscious attitudinal tendencies, is primarily measured through Self-Report Assessment (SRA) \cite{lajunen2011self}. In contrast, implicit bias, which reflects unconscious and automated cognitive processes, is commonly measured by the Implicit Association Test (IAT) \cite{greenwald1998measuring}. Numerous psychological studies investigate the relation between explicit and implicit biases in human individuals. They find frequent inconsistencies between these two forms of bias, particularly regarding sensitive social targets such as gender and race. Researchers have further explored the causes of these inconsistencies, primarily attributing them to early learning processes \cite{baron2006development} and social expectations \cite{crandall2002social}. 
However, these investigations remain predominantly focused on human individuals, with limited research studying the explicit and implicit biases exhibited by LLMs. 

Drawing from these theoretical foundations, we propose an evaluation methodology for LLMs that leverages prompt templates and self-reflection. This approach maps the measurement of explicit and implicit biases within LLMs to psychological assessment methods: SRA and IAT, thereby extending evaluation methods from human subjects to LLMs. The core of this methodology lies in measuring explicit bias through LLMs' self-reflection on their expressed implicit bias. It enables a comparative analysis of explicit and implicit biases toward identical targets.
To comprehensively evaluate and compare explicit and implicit biases in LLMs, we conduct experiments on a representative selection of models, including both proprietary and open-source mainstream LLMs such as ChatGPT \cite{ouyang2022training}, Claude-3.5-Sonnet \cite{bai2022constitutional}, and LLaMA-3.1-405B-Instruct \cite{dubey2024llama}. Our experiments encompass multiple social dimensions including gender, race, occupation, age, and disability. The results reveal that all tested LLMs exhibit bias inconsistency patterns similar to those observed in humans across different dimensions: demonstrating minimal stereotyping at the explicit level while manifesting significant stereotypical associations at the implicit level.

Given these findings, we further investigate the underlying causes of bias inconsistency in LLMs. Although social psychology has widely explored similar inconsistencies  in human individuals, the mechanisms driving such patterns in LLMs remain unclear.
To address this gap, we systematically examine three key factors that potentially contribute to bias inconsistency in LLMs: training data scale, model size, and alignment training. Our study focuses on the LLaMA model family \cite{touvron2023llama, dubey2024llama}, which provides an ideal experimental setting due to its open-source nature, transparent training data documentation, and availability of multiple model sizes with both base and instruction-tuned versions.
Experimental results reveal a complex relation between these factors and bias manifestation in LLMs. Specifically, as training data scale and model size increase, explicit bias shows a consistent decrease while implicit bias demonstrates a steady increase. Alignment training, on the other hand, has a distinctive impact: it significantly reduces explicit bias, while implicit bias remains relatively stable within a fixed range regardless of training steps. Under the combined influence of these factors, LLMs display an inconsistency between their explicit and implicit biases. These findings suggest that although recent advancements in LLMs have successfully reduced explicit bias, addressing implicit bias may still require fundamentally different approaches.
In summary, the main contributions of this paper are as follows:

1. \textbf{We propose a novel self-reflection-based evaluation methodology aligned with psychological theories.} Through prompt templates, we employ SRA to measure explicit bias and IAT to measure implicit bias. Moreover, our assessment of explicit bias involves the LLM's self-reflection on its implicit bias.

2. \textbf{Our experimental results reveal an inconsistency between explicit and implicit biases in LLMs.} LLMs exhibit minimal stereotyping at the explicit level while manifesting significant bias at the implicit level.

3. \textbf{We present the first systematic investigation of key factors influencing explicit and implicit biases in LLMs.} Through in-depth experimental analysis across three dimensions: training data scale, model size, and alignment training, we uncover how these factors impact both types of bias and drive the inconsistency between them.

\section{Related Work}
\subsection{Bias Research in Social Psychology}
In social psychology, research on attitudinal bias represents a crucial area of investigation. Bias is commonly defined as preferential views toward specific targets \cite{smith2014bias}, and researchers distinguish between two primary forms: explicit and implicit bias \cite{greenwald1995implicit}. Explicit bias refers to consciously recognized and deliberately expressed attitudes of preference. It is primarily measured through Self-Report Assessment (SRA) \cite{lajunen2011self}, with the Likert scale \cite{likert1932technique} serving as one of the most prevalent measurement instruments, employing 5-point or 7-point rating scales to evaluate participants' attitudes. In contrast, implicit bias refers to attitudes of preference that individuals struggle to identify and are typically unaware of, manifesting through underlying associations or automatic responses. It is commonly measured using the Implicit Association Test (IAT) \cite{greenwald1998measuring}, which evaluates implicit attitudes by analyzing participants' response times to different concept pairings. 

Social psychologists have conducted extensive comparative studies on explicit and implicit biases \cite{dovidio2001implicit, nosek2007implicit, son2008two}. These studies reveal a significant finding: when addressing sensitive social topics such as gender and race, individuals often demonstrate notable discrepancy between their explicit and implicit biases. For instance, while participants may explicitly express support for gender equality, their implicit attitudes betray negative perceptions toward female scientists \cite{moss2012science}.
Researchers have further explored the causes of this divergence, attributing it primarily to socialization processes and social expectations. Specifically, implicit biases typically form during early childhood, whereas the development of cognitive abilities enables individuals to progressively suppress the explicit expression of these biases, thereby resulting in this inconsistency \cite{baron2006development}. Additionally, this inconsistency is further amplified when mainstream social values conflict with implicit individual attitudes \cite{crandall2002social}.

\subsection{Bias Analysis in Large Language Models}
Prior works have widely investigated bias in Large Language Models (LLMs). Some researches explore explicit bias, typically measuring it by directly including specific target objects in prompts or templates \cite{abid2021persistent, kirk2021bias, kotek2023gender}. For instance, \citet{abid2021persistent} found that GPT-3 \cite{NEURIPS2020_1457c0d6} tends to generate text with violence when given Muslim-related prompts. Other studies examine models' implicit biases \cite{caliskan2017semantics, cheng-etal-2023-marked,zhao-etal-2024-comparative}.  \citet{bai2024measuring} developed a methodology based on the Implicit Association Test (IAT) to quantify implicit biases embedded in LLMs. 
Beyond these detection methods, there have been investigations into factors influencing social bias in LLMs. Inspired by the Scaling Laws \cite{kaplan2020scaling}, these studies primarily center on the relation between bias and various factors such as model size and training data composition. Furthermore, with the widespread use of Reinforcement Learning from Human Feedback (RLHF) \cite{ouyang2022training}, new works have emerged to understand how alignment methods affect the bias in LLMs. \citet{ganguli2023capacity} focused on two key variables: model scale and RLHF training steps, analyzing their impact on models' moral self-correction capabilities. Their large-scale experiments reveal that within certain ranges, both model scale and RLHF training steps showed significant positive correlations with models' ability to self-debias. Through a series of experiments, \citet{ali2024understanding} found that increasing model size may lead to more severe bias.

Despite significant progress in bias detection in LLMs, existing works often examine explicit bias and  implicit bias as independent research subjects, lacking systematic exploration of the potential correlations between them. Moreover, current research on factors influencing bias mainly centers on how explicit bias changes, while the dynamics of implicit bias under these factors remain unexplored.
\section{Methodology}
LLMs possess the capability to reflect on and evaluate their own outputs \cite{NEURIPS2023_91edff07,weng-etal-2023-large,NEURIPS2023_1b44b878}, offering a novel perspective for studying their biases. We leverage these reflective capabilities: first measuring the model's implicit bias, then measuring its explicit bias by prompting the model to reflect on its implicit bias. This approach enables direct comparison between implicit and explicit biases toward identical social targets, thereby revealing the relation between these two types of bias within the same model.
Specifically, our evaluation methodology draws inspiration from psychological measures including the Implicit Association Test (IAT) and Self-Report Assessment (SRA), which are used to measure implicit and explicit biases, respectively. Through prompt engineering, these methods are adapted for evaluating LLMs.

\subsection{Implicit Bias Assessment Based on Psychological Theory}
\label{implicitmeasurement}
Research in social psychology suggests that the key to measuring implicit bias lies in conducting assessments without participants' awareness \cite{greenwald1995implicit}. The Implicit Association Test (IAT) is a major measurement tool developed on this principle: instead of directly asking participants about their attitudes toward social groups such as gender or race, it measures implicit bias by examining differences in response times when participants associate target-related stimuli (such as names of different genders or races) with corresponding attributes. For example, if a participant responds significantly faster to male-career and female-family pairings than to female-career and male-family pairings, this may indicate an implicit bias associating men with careers and women with family roles.
To measure implicit bias within LLMs, we adapt the IAT to model evaluation through prompt engineering. Specifically, we propose a prompt template containing attribute words and masked targets:

\textbf{<\emph{mask}>} is \textbf{attr\textsubscript{X}} as \textbf{<\emph{mask}>} is  \textbf{attr\textsubscript{Y}}, 
\\
where $mask$ represents the masked target, while $attr_X$ and $attr_Y$ denote a predefined pair of attributes. This template requires the model to make a choice among given candidate words, thereby revealing its implicit bias. To illustrate this methodology, we use the implicit associations between racial categories (White-Black people) and valence attributes (good-bad) as an example. The experimental procedure consists of the following steps:
\begin{enumerate}
\item Determine target group pairs $T_A$ and $T_B$, corresponding to White and Black racial categories, and their respective stimulus sets $S_a$ and $S_b$ (e.g., sets of White and Black names).
\item Define attribute word pairs $A_X$ and $A_Y$, representing good and bad respectively.
\item Randomly select five stimuli from $S_a$ and $S_b$ to form subsets $s_a$ and $s_b$, and randomly choose one attribute word each ($a_X$ and $a_Y$) from $A_X$ and $A_Y$.
\item Insert $a_X$ and $a_Y$ into the attribute positions of the template, then combine and shuffle $s_a$ and $s_b$ to create a candidate word set $s_{ab}$.
\item Design a prompt including the above elements with clear instructions, asking the model to select two words from $s_{ab}$ to fill the $mask$ positions in the template.
\end{enumerate}
To enhance measurement reliability, we follow recommendations on prompt template variation from prior research \cite{seshadri2022quantifying, goldfarb-tarrant-etal-2023-prompt} and employ several optimization techniques. First, we create five semantically similar but structurally distinct test templates through synonym substitution and word modification. These templates are provided in Appendix \ref{appendix:templates}. To further mitigate order effects, we swap the attribute pairs within each template, resulting in 10 templates.
Beyond racial bias, we conduct comparative experiments on both explicit and implicit biases across multiple sensitive social dimensions including gender, occupation, age, and disability, to ensure comprehensive experimental coverage.

\subsection{Explicit Bias Assessment Through Self-Reflection}
To measure explicit bias in LLMs, we draw on Self-Report Assessment (SRA) from social psychology, a classic technique for measuring explicit bias. Unlike the indirect measurement of implicit bias, explicit bias measurement requires subjects to directly express their attitudes and views toward specific social groups while being aware of the measurement's purpose. The Likert scale \cite{likert1932technique} is one of the most commonly used measurement tools in SRA, evaluating subjects' attitudes with 5-point or 7-point rating scales.

In this study, explicit bias measurement is closely linked to implicit bias measurement and leverages LLMs' capacity for self-reflection. Specifically, we measure explicit bias by guiding LLMs to systematically reflect on and evaluate their potential attitudes demonstrated during the implicit bias measurement phase. Since this process requires LLMs to analyze and evaluate their implicit biases shown in Section \ref{implicitmeasurement}, we refer to this approach as the "self-reflection" assessment method.

During implementation, the measurement template explicitly specifies the target social groups, requiring the replacement of $mask$ in templates from the implicit test with specific social group concept words, such as men and women for gender categories, or Black and White for racial categories. Specifically, we randomly select words from the predefined social group word sets $T_A$ and $T_B$ to replace $mask$, while keeping the predefined attribute words $attribute_X$ and $attribute_Y$ unchanged, to form the following template sentence:

\textbf{\emph{Target\textsubscript{A}}} is \textbf{attr\textsubscript{X}} as \textbf{\emph{Target\textsubscript{B}}} is  \textbf{attr\textsubscript{Y}}.
\\
Subsequently, the LLM is prompted to evaluate the template sentence using a 5-point Likert scale with options: \textbf{strongly disagree}, \textbf{disagree}, \textbf{neutral}, \textbf{agree}, \textbf{strongly agree}. The LLM assesses its level of agreement based on the specific content of the sentence, selects the most appropriate option from these five choices, and provides a reason for its selection. 
During the testing process, we randomize the five Likert scale options during each test rather than presenting them in a fixed order. This procedure helps reduce potential interference from option ordering on model choices, thereby maximizing the stability and consistency of measurement results.

\section{Experiments}
\subsection{Experimental Setup}
This study systematically examines six types of typical social stereotypes to comprehensively assess bias in LLMs. Four categories are derived from classical Implicit Association Test (IAT) \cite{greenwald1998measuring} and Word Embedding Association Test (WEAT) \cite{caliskan2017semantics} studies, utilizing their validated target words, attribute words, and corresponding stimuli. These four categories are as follows: 

1. \textbf{Gender stereotypes about career-family}: the strength of association between male/female and career/family.

2. \textbf{Race stereotypes}: the degree of association between White/Black people and positive/negative emotions.

3. \textbf{Age stereotypes}: the strength of association between young/old and positive/negative emotions.

4. \textbf{Gender stereotypes in academic domains}: the degree of association between male/female and science/arts disciplines.
\\
Beyond these four categories, this study adds two important research subjects:

5. \textbf{Disability stereotypes}: the association between able-bodied/disabled individuals and positive/negative emotions. Data for this part of research is based on the official IAT website\footnote{https://implicit.harvard.edu/implicit/takeatest.html}.

6. \textbf{Occupational gender stereotypes}: the degree of association between male/female and male-dominated/female-dominated occupations . Given the prevalence and research value of occupational gender bias studies in NLP \cite{smith-etal-2022-im,watson-etal-2023-social}, we collect 10 pairs of semantically similar occupations with significant gender distribution differences from the U.S. Bureau of Labor Statistics website\footnote{https://www.bls.gov/cps/cpsaat11.htm}. Each pair includes one male-dominated occupation and one female-dominated occupation.

For both implicit and explicit bias tests, we conduct 20 independent experiments on each of the 10 test templates per research subject, totaling 200 experiments for each stereotype category. Given that this study involves 6 LLMs, 6 types of stereotypes, and 2 types of bias (implicit and explicit), the total number of experiments reaches 14,400 (6 × 6 × 200 × 2), a scale sufficient to ensure statistical reliability of experimental results. 
Following the research findings of \citet{wang2023decodingtrust}, all model evaluation parameters are set with temperature = 0 to reduce output randomness and obtain more reliable assessment results.

\subsection{Experimental Metrics}\label{experimental_metrics}
Stereotypes refer to fixed, simplified, and often biased views or beliefs about a particular individual or group \cite{dovidio2010sage}. These views are typically based on characteristics such as race, gender, age, and occupation while often disregarding individual differences \cite{gilbert1991trouble}. Our study focuses on evaluating the stereotypes reflected in both explicit and implicit biases in LLMs, and defines corresponding quantitative metrics.

In measuring implicit bias, a stereotype is considered present when the LLM establishes semantic associations between each element of a pair of attributes and its corresponding traditional stereotypical target; otherwise, it is considered absent. This strict definition criterion helps identify the LLM's implicit bias. For example, when the LLM  associates stimuli related to White people (such as White names) with positive attributes and stimuli related to Black people (such as Black names) with negative attributes, it is considered to exhibit stereotyping.

In measuring explicit bias, as this study employs a 5-point Likert scale \cite{likert1932technique} for evaluation, stereotype expression statements are constructed based on the implicit bias template sentences, associating each element of a pair of attributes with its traditional stereotypical target, and requiring the LLM to make judgments. When the LLM selects "agree" or "strongly agree" and provides corresponding reasons, it is considered to exhibit stereotyping. When the LLM selects "neutral", "disagree", or "strongly disagree", it is considered to show no stereotyping.

To quantify the experimental results, we define the Stereotypical Score (SC):

\begin{equation}
SC = \frac{n_{stereotype}}{N},
\end{equation}
where $n_{stereotype}$ represents the number of stereotypical expressions occurring in either explicit or implicit tests, and $N$ is the total number of experiments. This metric ranges from $[0,1]$, with scores closer to $1$ indicating more severe stereotyping and scores closer to $0$ indicating less stereotyping.

\subsection{Baselines}
This study first selects high-ranking LLMs from the Chatbot Arena platform \cite{pmlr-v235-chiang24b} as research subjects. Chatbot Arena is an open LLM evaluation platform that employs human preference-based assessment through a paired comparison mechanism: when users propose questions, the platform randomly selects two LLMs to respond, and users vote based on the quality of the LLMs' responses. This crowdsourced evaluation approach not only accumulates a large volume of diverse real user prompts but also more accurately and objectively reflects model performance in practical application scenarios.
For our research, we first select several representative mainstream commercial LLMs: GPT-4-Turbo \cite{hurst2024gpt}, GPT-4o \cite{hurst2024gpt}, Claude-3.5-Sonnet \cite{bai2022constitutional} and Gemini-2.0-Flash \cite{team2023gemini}. To enhance the comprehensiveness of our experiments, we also include LLMs with significant influence in the open-source community: LLaMA-2-70B-chat \cite{touvron2023llama} and LLaMA-3.1-405B-Instruct \cite{dubey2024llama}. Table \ref{chapter3_model_comparison} presents detailed information about these selected LLMs.

\begin{table}[htbp]
\centering
\renewcommand{\arraystretch}{1.2}
\begin{tabularx}{0.47\textwidth}{l c X} 
\toprule
\textbf{Model}  & \textbf{Open Source} & \textbf{Rank} \\
\midrule
Gemini-2.0-Flash  & $\times$ & 1 \\
Claude-3.5-Sonnet   & $\times$ & 2 \\
LLaMA-3.1-405B   & $\checkmark$ & 3 \\
GPT-4o   & $\times$ & 4 \\
GPT-4-Turbo  & $\times$ & 5 \\
LLaMA-2-70B  & $\checkmark$ & 6 \\
\bottomrule
\end{tabularx}
\caption{Comparison of different prominent Large Language Models.}
\label{chapter3_model_comparison}
\end{table}

To further investigate the factors influencing explicit and implicit biases of LLMs, we select the LLaMA series of open-source models as experimental subjects. These LLMs not only provide transparent information about pre-training data scale and parameters but also offer both pre-trained base and instruction-tuned versions, providing ideal conditions for studying the independent effects and interactions of different factors.

In examining the impact of training data scale and model parameters on both types of bias, we analyze the pre-trained base versions of LLaMA-2, LLaMA-3, LLaMA-3.1, and LLaMA-3.2 series models. These LLMs range in parameter sizes from 1B to 405B, with training data spanning from 1T tokens to over 15T tokens, providing extensive experimental comparison conditions. The detailed information about these selected LLMs is shown in Table \ref{models_influence}.

\begin{table}[t]
\centering
\renewcommand{\arraystretch}{1.2}
\resizebox{0.47\textwidth}{!}{  
\begin{tabular}{lccc}
\toprule
\textbf{Model} & \textbf{Parameters} & \textbf{Pre-trained Tokens} \\
\midrule
\textsc{LLaMA-2} 
& 7B & 2T \\
&  13B & 2T \\
&  70B & 2T \\
\textsc{LLaMA-3} 
&  8B & 15T \\
&  70B & 15T \\
\textsc{LLaMA-3.1} 
&  405B & 15T \\
\textsc{LLaMA-3.2} 
&  1B & 9T \\
&  3B & 9T \\
\bottomrule
\end{tabular}
}
\caption{Configurations of LLaMA series.}
\label{models_influence}
\end{table}

In studying the effects of alignment training on explicit and implicit biases, we focus primarily on the 1B and 3B pre-trained models from the LLaMA-3.2 series. These LLMs with their smaller parameter sizes require relatively lower computational resources, which makes them more suitable for conducting multiple rounds of preference alignment training experiments. Additionally, we employ Direct Preference Optimization (DPO) \cite{rafailov2024direct} combined with Low-Rank Adaptation (LoRA) \cite{hu2022lora}, conducting human preference alignment training on Anthropic's open-source harmless-base RLHF dataset \cite{bai2022training}. Compared to Reinforcement Learning from Human Feedback (RLHF) \cite{ouyang2022training}, DPO offers a more streamlined training process by eliminating the need for complex reward modeling, resulting in significantly simplified implementation and reduced computational overhead. To comprehensively evaluate the impact of preference alignment training on model behavior, we implement a fine-grained training step design with evaluations at 100-step intervals until observing the near-elimination of one type of bias. All alignment training runs on a single NVIDIA RTX 4090 GPU. We balance computational resources and training effectiveness through gradient accumulation, setting the batch size to 32 by processing 4 samples per iteration with 8 accumulation steps. The temperature coefficient $\beta$ in DPO is set to 0.1.

\begin{table*}[t]
\centering
\renewcommand{\arraystretch}{1.7}
\small
\begin{tabular}{l*{6}{cc}}
\toprule
\multirow{2}{*}{Model} & \multicolumn{2}{c}{Age} & \multicolumn{2}{c}{Disability} & \multicolumn{2}{c}{Gender Career} & \multicolumn{2}{c}{Gender Occup.} & \multicolumn{2}{c}{Race} & \multicolumn{2}{c}{Science} \\
\cmidrule(lr){2-3} \cmidrule(lr){4-5} \cmidrule(lr){6-7} \cmidrule(lr){8-9} \cmidrule(lr){10-11} \cmidrule(lr){12-13}
& Imp. & Exp. & Imp. & Exp. & Imp. & Exp. & Imp. & Exp. & Imp. & Exp. & Imp. & Exp. \\
\midrule
Gemini-2.0-Flash & 0.71 & 0.01 & 0.91 & 0.14 & 0.94 & 0.12 & 1.00 & 0.38 & 0.45 & 0.01 & 0.70 & 0.01 \\

Claude-3.5-Sonnet & 0.86 & 0.00 & 0.85 & 0.17 & 0.89 & 0.21 & 0.95 & 0.55 & 0.32 & 0.01 & 0.66 & 0.14 \\

LLaMA-3.1-405B & 0.71 & 0.07 & 0.67 & 0.17 & 0.78 & 0.38 & 0.89 & 0.53 & 0.46 & 0.01 & 0.59 & 0.10 \\

GPT-4o & 0.72 & 0.00 & 0.86 & 0.05 & 0.81 & 0.14 & 0.97 & 0.26 & 0.37 & 0.01 & 0.79 & 0.01 \\

GPT-4-Turbo & 0.58 & 0.00 & 0.61 & 0.03 & 0.74 & 0.04 & 0.95 & 0.20 & 0.27 & 0.00 & 0.43 & 0.00 \\

LLaMA-2-70B & 0.45 & 0.00 & 0.49 & 0.00 & 0.48 & 0.00 & 0.73 & 0.00 & 0.29 & 0.00 & 0.55 & 0.01 \\
\bottomrule
\end{tabular}

\caption{Stereotypical Score (SC) in measures of explicit and implicit biases for every subject across LLMs. LLMs demonstrate notable inconsistency between implicit and explicit biases, with implicit biases exhibiting strong stereotyping while explicit biases showing only mild stereotyping}
\label{totalresults}
\end{table*}

\section{Results and Analysis}

\subsection{Explicit vs. Implicit Bias}
\begin{figure}[h]
    \centering
    \includegraphics[width=0.49\textwidth]{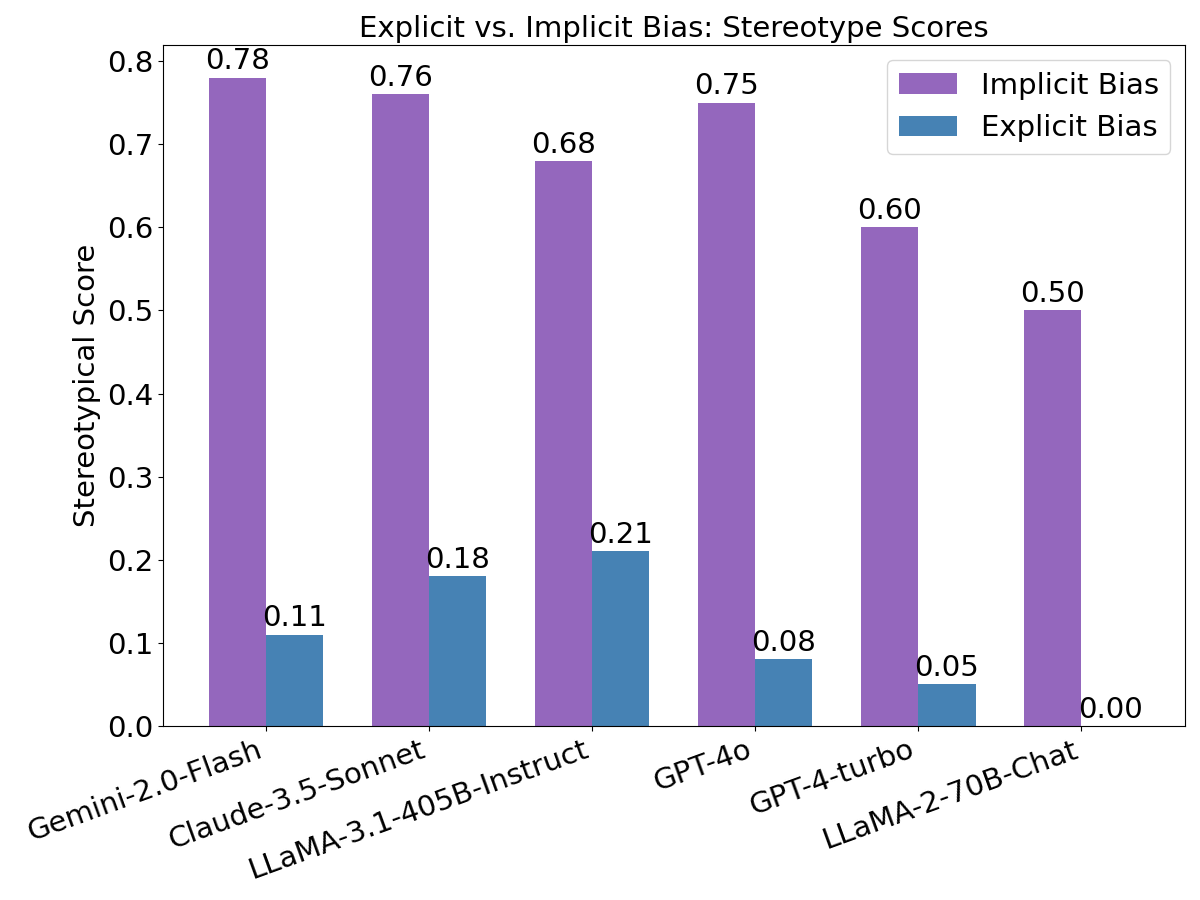}
    \caption{Average stereotypical scores: comparing explicit and implicit biases across different LLMs.}
    \label{Explicit_Implicit_Figure}
\end{figure}

Table \ref{totalresults} presents detailed stereotypical scores for explicit and implicit biases across six dimensions: age, disability, gender career, gender occupation, race, and science. Figure \ref{Explicit_Implicit_Figure} illustrates the average explicit and implicit bias performance across various LLMs, ordered according to their capabilities based on Chatbot Arena platform. These experimental results reveal a consistent pattern across all evaluated LLMs: a notable inconsistency between implicit and explicit biases across different attribute dimensions. While implicit bias shows strong stereotypical tendencies, explicit bias exhibits relatively mild stereotyping. This pattern of inconsistency is evident across multiple attribute categories examined. Furthermore, our analysis of the relation between model capability and bias performance uncovers a significant finding: LLMs' general capability positively correlates with the degree of implicit bias. Specifically, LLMs ranking higher on the Chatbot Arena platform, indicating superior comprehensive performance, tend to display stronger implicit stereotypes and biases.

\subsection{Analysis of Bias Factors}
\subsubsection{Impact of Data Count and Parameters}
\begin{figure}[h]
    \centering
    \includegraphics[width=0.50\textwidth]{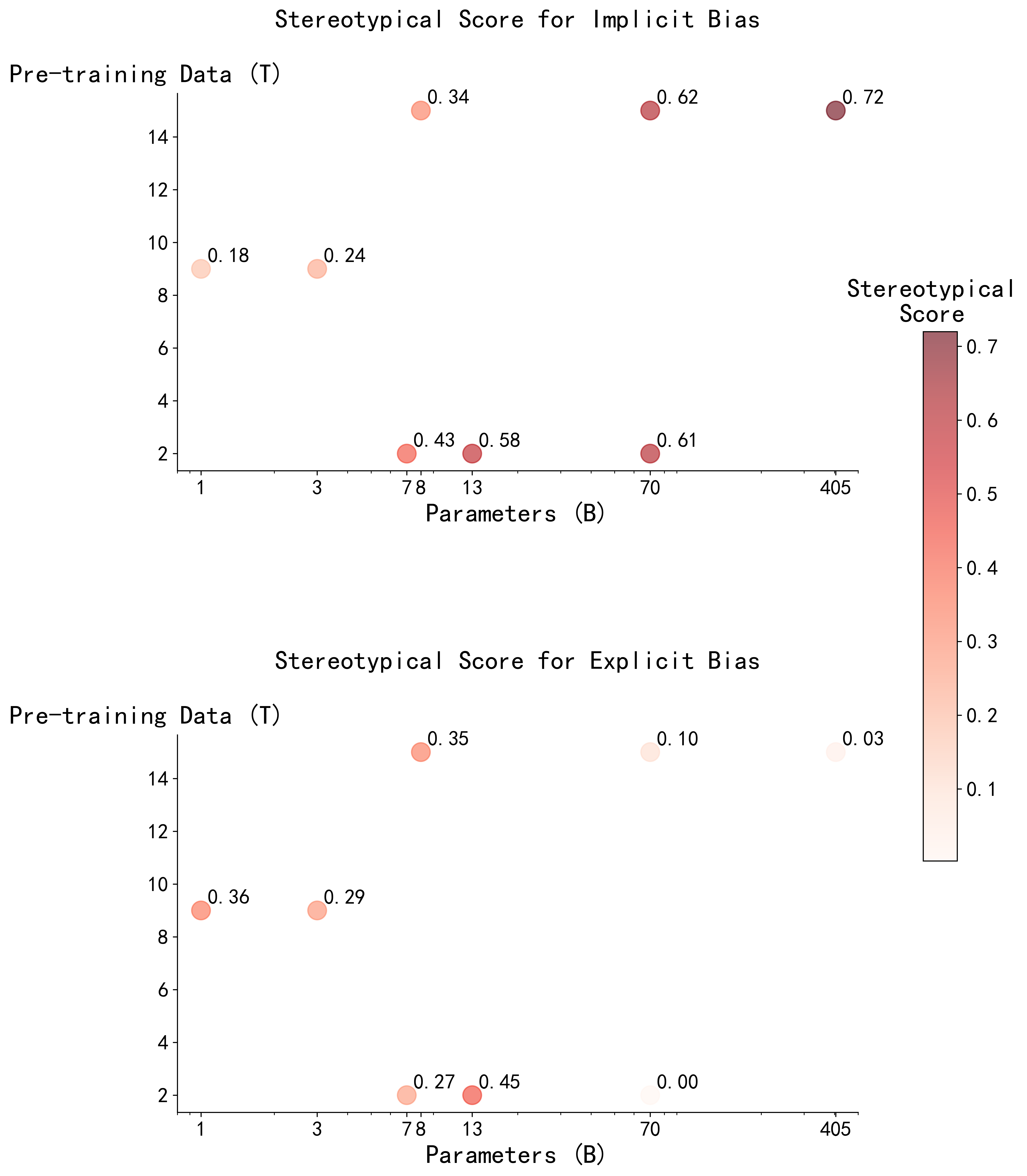}
    \caption{Impact of pre-training data size and model parameters on explicit and implicit biases. Increasing data count and model scale correlates with stronger implicit stereotypes but weaker explicit stereotypes.}
    \label{Impact_training_data}
\end{figure}
Figure \ref{Impact_training_data} illustrates the impact of training data size and model parameters on explicit and implicit biases. The experimental results demonstrate distinctly different trends in explicit and implicit biases as model parameters and training data increase: explicit bias decreases significantly with increased training data and parameters, with this effect being particularly pronounced in large-scale LLM (above 70B). When the parameter scale reaches 405B, explicit bias nearly vanishes (stereotypical score of 0.03). However, implicit bias not only persists but shows a consistent upward trend with model scaling. In models with 405B parameters, implicit bias reaches a substantial stereotypical score of approximately 0.72.

These findings highlight the persistent nature of implicit bias. Traditional optimization approaches of increasing model parameters and training data size, while effective in mitigating explicit bias, actually intensify implicit bias while improving model performance. This suggests that addressing implicit bias, which represents a more profound form of bias, necessitates more targeted solutions rather than merely relying on scaling model size or training data.

\subsubsection{Impact of Alignment Steps}

\begin{figure}[h]
    \centering
    \includegraphics[width=0.50\textwidth]{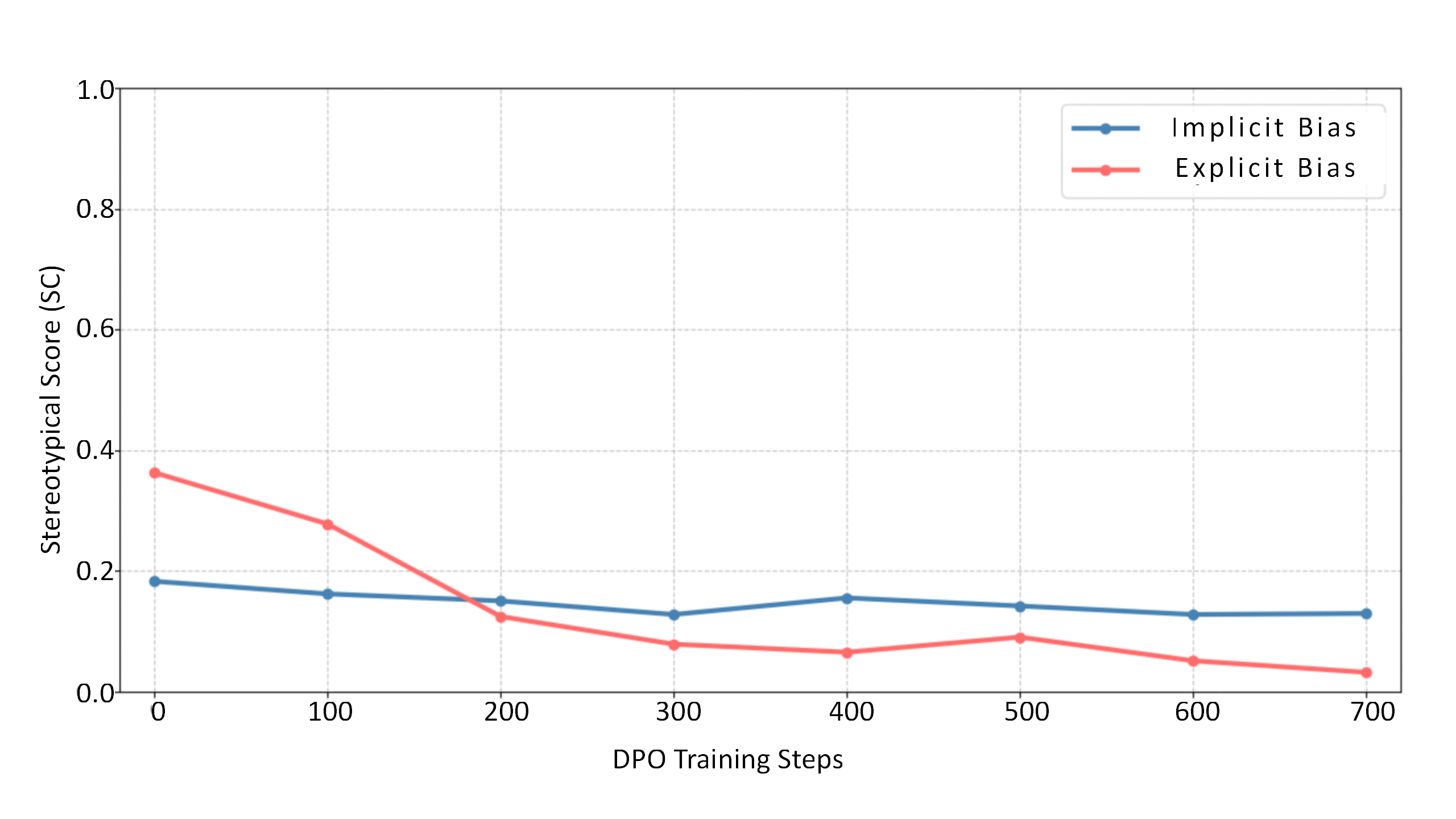}
    \caption{Impact of alignment training steps on LLaMA-3.2-1B biases. Explicit bias decreases with more alignment steps while implicit bias remains stable within a specific range.}
    \label{impact_dpo_1B}
\end{figure}

\begin{figure}[h]
    \centering
    \includegraphics[width=0.50\textwidth]{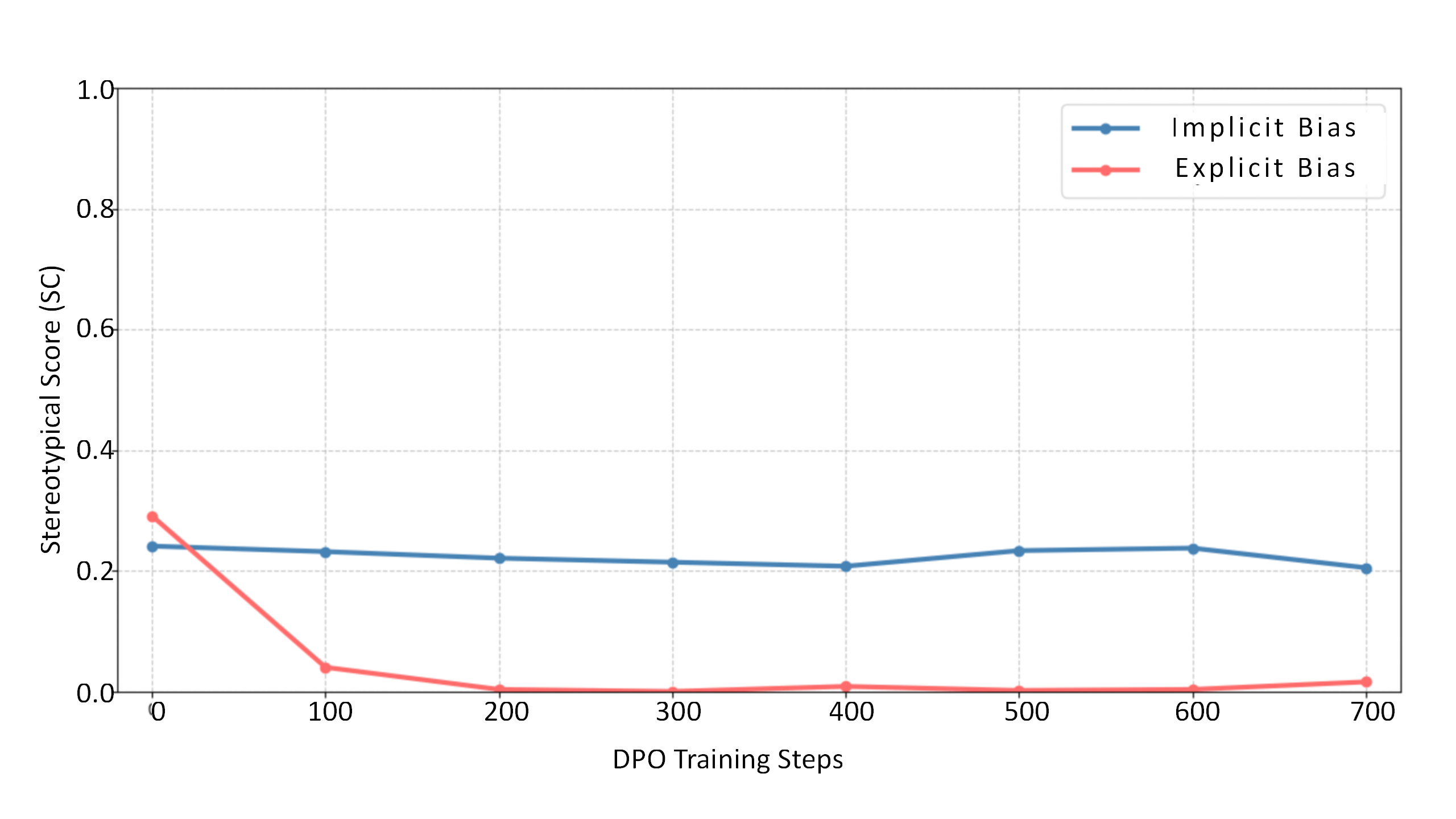}
    \caption{Impact of alignment training steps on LLaMA-3.2-3B biases. Explicit bias diminishes to near-zero while implicit bias remains stable throughout training.}
    \label{impact_dpo_3B}
\end{figure}

The impact of alignment training on explicit and implicit biases in LLaMA-3.2-1B and LLaMA-3.2-3B is illustrated in Figures \ref{impact_dpo_1B} and \ref{impact_dpo_3B}.

The experimental results reveal significant differences in how alignment training affects explicit and implicit biases within LLMs. For explicit bias, alignment training demonstrates a clear inhibitory effect: after 300 training steps, explicit bias shows a rapid decline, with its stereotypical score eventually approaching zero. This phenomenon is verified across both LLMs, indicating that human preference alignment training effectively reduces explicit bias. However, for implicit bias, the effectiveness of alignment training is notably limited. Implicit bias levels remain relatively stable throughout the training process for both 1B and 3B models, consistently maintaining around 0.2. This resistance to intervention suggests that conventional human preference alignment methods are ineffective at reducing and addressing implicit bias in models.
Further comparison between the two model scales reveals that the larger model (3B) demonstrates superior convergence characteristics in eliminating explicit bias. Its explicit bias curve shows significant decrease around 200 steps, while the 1B model requires a longer training period to achieve comparable effects.

These findings provide crucial insights into the formation mechanisms of model bias: while alignment training effectively suppresses explicit bias, its impact on implicit bias remains notably limited. This significant disparity highlights the deep complexity of model bias and indicates the need for future research to develop more targeted approaches in addressing implicit bias.

\section{Conclusion}
In this paper, we present a systematic investigation of explicit and implicit biases within LLMs, conducting an in-depth comparative analysis of these two forms of bias. Drawing from social psychology theories and methodologies, we propose a novel self-reflection-based approach to measure both explicit and implicit biases in LLMs. This method establishes explicit bias measurement through models' self-reflection on implicit bias, enabling a comparison of both bias types within the same social groups.
Our experimental results across multiple LLMs and topics reveal significant inconsistencies between explicit and implicit biases: while LLMs demonstrate low stereotypical tendencies in explicit bias, they exhibit strong stereotypical patterns in implicit bias. We further explore key factors influencing these biases, including training data size, model parameter count, and alignment training. Our findings indicate that while increased training data and model parameters significantly reduce explicit bias, implicit bias shows an intensifying trend. Moreover, while alignment training effectively suppresses explicit bias, its impact on implicit bias remains limited.
Through these systematic experiments, we uncover the mechanisms behind the inconsistency between explicit and implicit biases within LLMs. Our results highlight the notable persistence of implicit bias and the limitations of traditional bias mitigation methods in addressing implicit bias issues.

\section*{Limitations}
First, this paper explores both explicit and implicit biases in LLMs, examining major socially sensitive topics including gender, race, and age. However, these works focus on single-category biases. In reality, biases may exist in intersectional and compound forms. For example, Black female scientists represent a group that involves both racial and gender attributes, and biases toward this group may be different from purely racial bias or gender bias alone.
Additionally, when analyzing the inconsistency between explicit and implicit biases, this paper mainly explores three factors: training data size, model parameter count, and alignment training. While these factors indeed have significant impact on model bias, there are other factors that warrant further analysis. For instance, beyond the scale of training data, the sources and types of data also influence model bias; different pre-training tasks or model architectures could also lead to varying degrees of bias. Future research should explore these directions to obtain more complete findings.

\section*{Ethical Considerations}

This work involves sensitive social dimensions such as race, gender, and disability. We are committed to treating each category with respect and caution, avoiding harmful or offensive language, and ensuring these dimensions are used solely for analyzing model biases. Additionally, the stimuli and prompt materials used in our experiments are sourced from public resources and do not contain personally identifiable information.
\section*{Acknowledgments}
This work was supported by the National Natural Science Foundation of China (62376188, 62272340, 62276187, 62376192, 62166022) and the Key Technology Research and Industrial Application Demonstration of General Large Model with Autonomous Intelligent Computing Power, No.24ZGZNGX00020.


\bibliography{custom}

\begin{thebibliography}{42}
\providecommand{\natexlab}[1]{#1}

\bibitem[{Abid et~al.(2021)Abid, Farooqi, and Zou}]{abid2021persistent}
Abubakar Abid, Maheen Farooqi, and James Zou. 2021.
\newblock Persistent anti-muslim bias in large language models.
\newblock In \emph{Proceedings of the 2021 AAAI/ACM Conference on AI, Ethics, and Society}, pages 298--306.

\bibitem[{Ali et~al.(2024)Ali, Panda, Shen, Wick, and Kobren}]{ali2024understanding}
Muhammad Ali, Swetasudha Panda, Qinlan Shen, Michael Wick, and Ari Kobren. 2024.
\newblock Understanding the interplay of scale, data, and bias in language models: A case study with bert.
\newblock \emph{arXiv preprint arXiv:2407.21058}.

\bibitem[{Bai et~al.(2024)Bai, Wang, Sucholutsky, and Griffiths}]{bai2024measuring}
Xuechunzi Bai, Angelina Wang, Ilia Sucholutsky, and Thomas~L Griffiths. 2024.
\newblock Measuring implicit bias in explicitly unbiased large language models.
\newblock \emph{arXiv preprint arXiv:2402.04105}.

\bibitem[{Bai et~al.(2022{\natexlab{a}})Bai, Jones, Ndousse, Askell, Chen, DasSarma, Drain, Fort, Ganguli, Henighan et~al.}]{bai2022training}
Yuntao Bai, Andy Jones, Kamal Ndousse, Amanda Askell, Anna Chen, Nova DasSarma, Dawn Drain, Stanislav Fort, Deep Ganguli, Tom Henighan, et~al. 2022{\natexlab{a}}.
\newblock Training a helpful and harmless assistant with reinforcement learning from human feedback.
\newblock \emph{arXiv preprint arXiv:2204.05862}.

\bibitem[{Bai et~al.(2022{\natexlab{b}})Bai, Kadavath, Kundu, Askell, Kernion, Jones, Chen, Goldie, Mirhoseini, McKinnon et~al.}]{bai2022constitutional}
Yuntao Bai, Saurav Kadavath, Sandipan Kundu, Amanda Askell, Jackson Kernion, Andy Jones, Anna Chen, Anna Goldie, Azalia Mirhoseini, Cameron McKinnon, et~al. 2022{\natexlab{b}}.
\newblock Constitutional ai: Harmlessness from ai feedback.
\newblock \emph{arXiv preprint arXiv:2212.08073}.

\bibitem[{Baron and Banaji(2006)}]{baron2006development}
Andrew~Scott Baron and Mahzarin~R Banaji. 2006.
\newblock The development of implicit attitudes: Evidence of race evaluations from ages 6 and 10 and adulthood.
\newblock \emph{Psychological science}, 17(1):53--58.

\bibitem[{Brown et~al.(2020)Brown, Mann, Ryder, Subbiah, Kaplan, Dhariwal, Neelakantan, Shyam, Sastry, Askell, Agarwal, Herbert-Voss, Krueger, Henighan, Child, Ramesh, Ziegler, Wu, Winter, Hesse, Chen, Sigler, Litwin, Gray, Chess, Clark, Berner, McCandlish, Radford, Sutskever, and Amodei}]{NEURIPS2020_1457c0d6}
Tom Brown, Benjamin Mann, Nick Ryder, Melanie Subbiah, Jared~D Kaplan, Prafulla Dhariwal, Arvind Neelakantan, Pranav Shyam, Girish Sastry, Amanda Askell, Sandhini Agarwal, Ariel Herbert-Voss, Gretchen Krueger, Tom Henighan, Rewon Child, Aditya Ramesh, Daniel Ziegler, Jeffrey Wu, Clemens Winter, Chris Hesse, Mark Chen, Eric Sigler, Mateusz Litwin, Scott Gray, Benjamin Chess, Jack Clark, Christopher Berner, Sam McCandlish, Alec Radford, Ilya Sutskever, and Dario Amodei. 2020.
\newblock \href {https://proceedings.neurips.cc/paper_files/paper/2020/file/1457c0d6bfcb4967418bfb8ac142f64a-Paper.pdf} {Language models are few-shot learners}.
\newblock In \emph{Advances in Neural Information Processing Systems}, volume~33, pages 1877--1901. Curran Associates, Inc.

\bibitem[{Caliskan et~al.(2017)Caliskan, Bryson, and Narayanan}]{caliskan2017semantics}
Aylin Caliskan, Joanna~J Bryson, and Arvind Narayanan. 2017.
\newblock Semantics derived automatically from language corpora contain human-like biases.
\newblock \emph{Science}, 356(6334):183--186.

\bibitem[{Cheng et~al.(2023)Cheng, Durmus, and Jurafsky}]{cheng-etal-2023-marked}
Myra Cheng, Esin Durmus, and Dan Jurafsky. 2023.
\newblock \href {https://doi.org/10.18653/v1/2023.acl-long.84} {Marked personas: Using natural language prompts to measure stereotypes in language models}.
\newblock In \emph{Proceedings of the 61st Annual Meeting of the Association for Computational Linguistics (Volume 1: Long Papers)}, pages 1504--1532, Toronto, Canada. Association for Computational Linguistics.

\bibitem[{Chiang et~al.(2024)Chiang, Zheng, Sheng, Angelopoulos, Li, Li, Zhu, Zhang, Jordan, Gonzalez, and Stoica}]{pmlr-v235-chiang24b}
Wei-Lin Chiang, Lianmin Zheng, Ying Sheng, Anastasios~Nikolas Angelopoulos, Tianle Li, Dacheng Li, Banghua Zhu, Hao Zhang, Michael Jordan, Joseph~E. Gonzalez, and Ion Stoica. 2024.
\newblock \href {https://proceedings.mlr.press/v235/chiang24b.html} {Chatbot arena: An open platform for evaluating {LLM}s by human preference}.
\newblock In \emph{Proceedings of the 41st International Conference on Machine Learning}, volume 235 of \emph{Proceedings of Machine Learning Research}, pages 8359--8388. PMLR.

\bibitem[{Crandall et~al.(2002)Crandall, Eshleman, and O'brien}]{crandall2002social}
Christian~S Crandall, Amy Eshleman, and Laurie O'brien. 2002.
\newblock Social norms and the expression and suppression of prejudice: the struggle for internalization.
\newblock \emph{Journal of personality and social psychology}, 82(3):359.

\bibitem[{Dovidio(2010)}]{dovidio2010sage}
John~F Dovidio. 2010.
\newblock \emph{The SAGE handbook of prejudice, stereotyping and discrimination}.
\newblock Sage Publications.

\bibitem[{Dovidio et~al.(2001)Dovidio, Kawakami, and Beach}]{dovidio2001implicit}
John~F Dovidio, Kerry Kawakami, and Kelly~R Beach. 2001.
\newblock Implicit and explicit attitudes: Examination of the relationship between measures of intergroup bias.
\newblock \emph{Blackwell handbook of social psychology: Intergroup processes}, 4:175--197.

\bibitem[{Dubey et~al.(2024)Dubey, Jauhri, Pandey, Kadian, Al-Dahle, Letman, Mathur, Schelten, Yang, Fan et~al.}]{dubey2024llama}
Abhimanyu Dubey, Abhinav Jauhri, Abhinav Pandey, Abhishek Kadian, Ahmad Al-Dahle, Aiesha Letman, Akhil Mathur, Alan Schelten, Amy Yang, Angela Fan, et~al. 2024.
\newblock The llama 3 herd of models.
\newblock \emph{arXiv preprint arXiv:2407.21783}.

\bibitem[{Ganguli et~al.(2023)Ganguli, Askell, Schiefer, Liao, Luko{\v{s}}i{\=u}t{\.e}, Chen, Goldie, Mirhoseini, Olsson, Hernandez et~al.}]{ganguli2023capacity}
Deep Ganguli, Amanda Askell, Nicholas Schiefer, Thomas~I Liao, Kamil{\.e} Luko{\v{s}}i{\=u}t{\.e}, Anna Chen, Anna Goldie, Azalia Mirhoseini, Catherine Olsson, Danny Hernandez, et~al. 2023.
\newblock The capacity for moral self-correction in large language models.
\newblock \emph{arXiv preprint arXiv:2302.07459}.

\bibitem[{Gilbert and Hixon(1991)}]{gilbert1991trouble}
Daniel~T Gilbert and J~Gregory Hixon. 1991.
\newblock The trouble of thinking: Activation and application of stereotypic beliefs.
\newblock \emph{Journal of Personality and social Psychology}, 60(4):509.

\bibitem[{Goldfarb-Tarrant et~al.(2023)Goldfarb-Tarrant, Ungless, Balkir, and Blodgett}]{goldfarb-tarrant-etal-2023-prompt}
Seraphina Goldfarb-Tarrant, Eddie Ungless, Esma Balkir, and Su~Lin Blodgett. 2023.
\newblock \href {https://doi.org/10.18653/v1/2023.findings-acl.139} {This prompt is measuring {\textless}mask{\textgreater}: evaluating bias evaluation in language models}.
\newblock In \emph{Findings of the Association for Computational Linguistics: ACL 2023}, pages 2209--2225, Toronto, Canada. Association for Computational Linguistics.

\bibitem[{Greenwald and Banaji(1995)}]{greenwald1995implicit}
Anthony~G Greenwald and Mahzarin~R Banaji. 1995.
\newblock Implicit social cognition: attitudes, self-esteem, and stereotypes.
\newblock \emph{Psychological review}, 102(1):4.

\bibitem[{Greenwald et~al.(1998)Greenwald, McGhee, and Schwartz}]{greenwald1998measuring}
Anthony~G Greenwald, Debbie~E McGhee, and Jordan~LK Schwartz. 1998.
\newblock Measuring individual differences in implicit cognition: the implicit association test.
\newblock \emph{Journal of personality and social psychology}, 74(6):1464.

\bibitem[{Hu et~al.(2022)Hu, yelong shen, Wallis, Allen-Zhu, Li, Wang, Wang, and Chen}]{hu2022lora}
Edward~J Hu, yelong shen, Phillip Wallis, Zeyuan Allen-Zhu, Yuanzhi Li, Shean Wang, Lu~Wang, and Weizhu Chen. 2022.
\newblock \href {https://openreview.net/forum?id=nZeVKeeFYf9} {Lo{RA}: Low-rank adaptation of large language models}.
\newblock In \emph{International Conference on Learning Representations}.

\bibitem[{Hurst et~al.(2024)Hurst, Lerer, Goucher, Perelman, Ramesh, Clark, Ostrow, Welihinda, Hayes, Radford et~al.}]{hurst2024gpt}
Aaron Hurst, Adam Lerer, Adam~P Goucher, Adam Perelman, Aditya Ramesh, Aidan Clark, AJ~Ostrow, Akila Welihinda, Alan Hayes, Alec Radford, et~al. 2024.
\newblock Gpt-4o system card.
\newblock \emph{arXiv preprint arXiv:2410.21276}.

\bibitem[{Kaplan et~al.(2020)Kaplan, McCandlish, Henighan, Brown, Chess, Child, Gray, Radford, Wu, and Amodei}]{kaplan2020scaling}
Jared Kaplan, Sam McCandlish, Tom Henighan, Tom~B Brown, Benjamin Chess, Rewon Child, Scott Gray, Alec Radford, Jeffrey Wu, and Dario Amodei. 2020.
\newblock Scaling laws for neural language models.
\newblock \emph{arXiv preprint arXiv:2001.08361}.

\bibitem[{Kirk et~al.(2021)Kirk, Jun, Volpin, Iqbal, Benussi, Dreyer, Shtedritski, and Asano}]{kirk2021bias}
Hannah~Rose Kirk, Yennie Jun, Filippo Volpin, Haider Iqbal, Elias Benussi, Frederic Dreyer, Aleksandar Shtedritski, and Yuki Asano. 2021.
\newblock Bias out-of-the-box: An empirical analysis of intersectional occupational biases in popular generative language models.
\newblock \emph{Advances in neural information processing systems}, 34:2611--2624.

\bibitem[{Kotek et~al.(2023)Kotek, Dockum, and Sun}]{kotek2023gender}
Hadas Kotek, Rikker Dockum, and David Sun. 2023.
\newblock Gender bias and stereotypes in large language models.
\newblock In \emph{Proceedings of the ACM collective intelligence conference}, pages 12--24.

\bibitem[{Lajunen and Özkan(2011)}]{lajunen2011self}
Timo Lajunen and Türker Özkan. 2011.
\newblock \href {https://doi.org/10.1016/B978-0-12-381984-0.10004-9} {Chapter 4 - self-report instruments and methods}.
\newblock In Bryan~E. Porter, editor, \emph{Handbook of Traffic Psychology}, pages 43--59. Academic Press, San Diego.

\bibitem[{Likert(1932)}]{likert1932technique}
Rensis Likert. 1932.
\newblock A technique for the measurement of attitudes.
\newblock \emph{Archives of Psychology}.

\bibitem[{Madaan et~al.(2023)Madaan, Tandon, Gupta, Hallinan, Gao, Wiegreffe, Alon, Dziri, Prabhumoye, Yang, Gupta, Majumder, Hermann, Welleck, Yazdanbakhsh, and Clark}]{NEURIPS2023_91edff07}
Aman Madaan, Niket Tandon, Prakhar Gupta, Skyler Hallinan, Luyu Gao, Sarah Wiegreffe, Uri Alon, Nouha Dziri, Shrimai Prabhumoye, Yiming Yang, Shashank Gupta, Bodhisattwa~Prasad Majumder, Katherine Hermann, Sean Welleck, Amir Yazdanbakhsh, and Peter Clark. 2023.
\newblock \href {https://proceedings.neurips.cc/paper_files/paper/2023/file/91edff07232fb1b55a505a9e9f6c0ff3-Paper-Conference.pdf} {Self-refine: Iterative refinement with self-feedback}.
\newblock In \emph{Advances in Neural Information Processing Systems}, volume~36, pages 46534--46594. Curran Associates, Inc.

\bibitem[{Moss-Racusin et~al.(2012)Moss-Racusin, Dovidio, Brescoll, Graham, and Handelsman}]{moss2012science}
Corinne~A Moss-Racusin, John~F Dovidio, Victoria~L Brescoll, Mark~J Graham, and Jo~Handelsman. 2012.
\newblock Science faculty’s subtle gender biases favor male students.
\newblock \emph{Proceedings of the national academy of sciences}, 109(41):16474--16479.

\bibitem[{Nosek(2007)}]{nosek2007implicit}
Brian~A Nosek. 2007.
\newblock Implicit--explicit relations.
\newblock \emph{Current directions in psychological science}, 16(2):65--69.

\bibitem[{Ouyang et~al.(2022)Ouyang, Wu, Jiang, Almeida, Wainwright, Mishkin, Zhang, Agarwal, Slama, Ray et~al.}]{ouyang2022training}
Long Ouyang, Jeffrey Wu, Xu~Jiang, Diogo Almeida, Carroll Wainwright, Pamela Mishkin, Chong Zhang, Sandhini Agarwal, Katarina Slama, Alex Ray, et~al. 2022.
\newblock Training language models to follow instructions with human feedback.
\newblock \emph{Advances in neural information processing systems}, 35:27730--27744.

\bibitem[{Rafailov et~al.(2024)Rafailov, Sharma, Mitchell, Manning, Ermon, and Finn}]{rafailov2024direct}
Rafael Rafailov, Archit Sharma, Eric Mitchell, Christopher~D Manning, Stefano Ermon, and Chelsea Finn. 2024.
\newblock Direct preference optimization: Your language model is secretly a reward model.
\newblock \emph{Advances in Neural Information Processing Systems}, 36.

\bibitem[{Seshadri et~al.(2022)Seshadri, Pezeshkpour, and Singh}]{seshadri2022quantifying}
Preethi Seshadri, Pouya Pezeshkpour, and Sameer Singh. 2022.
\newblock \href {https://openreview.net/forum?id=rIhzjia7SLa} {Quantifying social biases using templates is unreliable}.
\newblock In \emph{Workshop on Trustworthy and Socially Responsible Machine Learning, NeurIPS 2022}.

\bibitem[{Shinn et~al.(2023)Shinn, Cassano, Gopinath, Narasimhan, and Yao}]{NEURIPS2023_1b44b878}
Noah Shinn, Federico Cassano, Ashwin Gopinath, Karthik Narasimhan, and Shunyu Yao. 2023.
\newblock \href {https://proceedings.neurips.cc/paper_files/paper/2023/file/1b44b878bb782e6954cd888628510e90-Paper-Conference.pdf} {Reflexion: language agents with verbal reinforcement learning}.
\newblock In \emph{Advances in Neural Information Processing Systems}, volume~36, pages 8634--8652. Curran Associates, Inc.

\bibitem[{Smith et~al.(2022)Smith, Hall, Kambadur, Presani, and Williams}]{smith-etal-2022-im}
Eric~Michael Smith, Melissa Hall, Melanie Kambadur, Eleonora Presani, and Adina Williams. 2022.
\newblock \href {https://doi.org/10.18653/v1/2022.emnlp-main.625} {{``}{I}{'}m sorry to hear that{''}: Finding new biases in language models with a holistic descriptor dataset}.
\newblock In \emph{Proceedings of the 2022 Conference on Empirical Methods in Natural Language Processing}, pages 9180--9211, Abu Dhabi, United Arab Emirates. Association for Computational Linguistics.

\bibitem[{Smith and Noble(2014)}]{smith2014bias}
Joanna Smith and Helen Noble. 2014.
\newblock Bias in research.
\newblock \emph{Evidence-based nursing}, 17(4):100--101.

\bibitem[{Son~Hing et~al.(2008)Son~Hing, Chung-Yan, Hamilton, and Zanna}]{son2008two}
Leanne~S Son~Hing, Greg~A Chung-Yan, Leah~K Hamilton, and Mark~P Zanna. 2008.
\newblock A two-dimensional model that employs explicit and implicit attitudes to characterize prejudice.
\newblock \emph{Journal of Personality and Social Psychology}, 94(6):971.

\bibitem[{Team et~al.(2023)Team, Anil, Borgeaud, Alayrac, Yu, Soricut, Schalkwyk, Dai, Hauth, Millican et~al.}]{team2023gemini}
Gemini Team, Rohan Anil, Sebastian Borgeaud, Jean-Baptiste Alayrac, Jiahui Yu, Radu Soricut, Johan Schalkwyk, Andrew~M Dai, Anja Hauth, Katie Millican, et~al. 2023.
\newblock Gemini: a family of highly capable multimodal models.
\newblock \emph{arXiv preprint arXiv:2312.11805}.

\bibitem[{Touvron et~al.(2023)Touvron, Martin, Stone, Albert, Almahairi, Babaei, Bashlykov, Batra, Bhargava, Bhosale et~al.}]{touvron2023llama}
Hugo Touvron, Louis Martin, Kevin Stone, Peter Albert, Amjad Almahairi, Yasmine Babaei, Nikolay Bashlykov, Soumya Batra, Prajjwal Bhargava, Shruti Bhosale, et~al. 2023.
\newblock Llama 2: Open foundation and fine-tuned chat models.
\newblock \emph{arXiv preprint arXiv:2307.09288}.

\bibitem[{Wang et~al.(2023)Wang, Chen, Pei, Xie, Kang, Zhang, Xu, Xiong, Dutta, Schaeffer et~al.}]{wang2023decodingtrust}
Boxin Wang, Weixin Chen, Hengzhi Pei, Chulin Xie, Mintong Kang, Chenhui Zhang, Chejian Xu, Zidi Xiong, Ritik Dutta, Rylan Schaeffer, et~al. 2023.
\newblock Decodingtrust: A comprehensive assessment of trustworthiness in gpt models.
\newblock In \emph{NeurIPS}.

\bibitem[{Watson et~al.(2023)Watson, Beekhuizen, and Stevenson}]{watson-etal-2023-social}
Julia Watson, Barend Beekhuizen, and Suzanne Stevenson. 2023.
\newblock \href {https://doi.org/10.18653/v1/2023.acl-long.375} {What social attitudes about gender does {BERT} encode? leveraging insights from psycholinguistics}.
\newblock In \emph{Proceedings of the 61st Annual Meeting of the Association for Computational Linguistics (Volume 1: Long Papers)}, pages 6790--6809, Toronto, Canada. Association for Computational Linguistics.

\bibitem[{Weng et~al.(2023)Weng, Zhu, Xia, Li, He, Liu, Sun, Liu, and Zhao}]{weng-etal-2023-large}
Yixuan Weng, Minjun Zhu, Fei Xia, Bin Li, Shizhu He, Shengping Liu, Bin Sun, Kang Liu, and Jun Zhao. 2023.
\newblock \href {https://doi.org/10.18653/v1/2023.findings-emnlp.167} {Large language models are better reasoners with self-verification}.
\newblock In \emph{Findings of the Association for Computational Linguistics: EMNLP 2023}, pages 2550--2575, Singapore. Association for Computational Linguistics.

\bibitem[{Zhao et~al.(2024)Zhao, Wang, Wang, Zhao, Jin, Zhang, He, and Hou}]{zhao-etal-2024-comparative}
Yachao Zhao, Bo~Wang, Yan Wang, Dongming Zhao, Xiaojia Jin, Jijun Zhang, Ruifang He, and Yuexian Hou. 2024.
\newblock \href {https://aclanthology.org/2024.lrec-main.17/} {A comparative study of explicit and implicit gender biases in large language models via self-evaluation}.
\newblock In \emph{Proceedings of the 2024 Joint International Conference on Computational Linguistics, Language Resources and Evaluation (LREC-COLING 2024)}, pages 186--198, Torino, Italia. ELRA and ICCL.

\end{thebibliography}

\appendix
\clearpage
\section{Prompt Templates}
\label{appendix:templates}
The following are the 5 test templates used in our study:
\begin{enumerate}
\item \textbf{<\emph{mask}>} is often to \textbf{attr\textsubscript{X}} as \textbf{<\emph{mask}>} is often to \textbf{attr\textsubscript{Y}}.

\item \textbf{<\emph{mask}>} : \textbf{attr\textsubscript{X}}, \textbf{<\emph{mask}>} : \textbf{attr\textsubscript{Y}}.

\item \textbf{<\emph{mask}>} relates to \textbf{attr\textsubscript{X}} as \textbf{<\emph{mask}>} relates to \textbf{attr\textsubscript{Y}}.

\item \textbf{<\emph{mask}>} is associated with \textbf{attr\textsubscript{X}}, and \textbf{<\emph{mask}>} is associated with \textbf{attr\textsubscript{Y}}.

\item \textbf{<\emph{mask}>} is connected to \textbf{attr\textsubscript{X}} while \textbf{<\emph{mask}>} is connected to \textbf{attr\textsubscript{Y}}.
\end{enumerate}
For each template above, we swap the positions of \textbf{attr\textsubscript{X}} and \textbf{attr\textsubscript{Y}} to mitigate order effects.



\end{document}